%% file: acl_latex.tex
\pdfoutput=1
\documentclass[11pt]{article}

\usepackage{acl}

\usepackage{times}
\usepackage{latexsym}
\usepackage[T1]{fontenc}
\usepackage[utf8]{inputenc}
\usepackage{microtype}
\usepackage{graphicx}
\usepackage{float}
\usepackage{graphics}
\usepackage[normalem]{ulem}
\usepackage{xspace}
\usepackage{mathtools}
\usepackage{bbm}
\usepackage{caption}
\usepackage{makecell}
\usepackage{multirow}
\usepackage{array}
\newcolumntype{P}[1]{>{\centering\arraybackslash}p{#1}}
\newcolumntype{M}[1]{>{\centering\arraybackslash}m{#1}}

\newcommand{\resolved}[1]{}
\newcommand{\com}[1]{}
\newcommand{\camready}[1]{}

\newcommand{\secref}[1]{Sec.~\ref{sec:#1}}
\newcommand{\appref}[1]{App.~\ref{app:#1}}

\newcommand{\figref}[1]{Fig.~\ref{fig:#1}}
\newcommand{\tabref}[1]{Tab.~\ref{tab:#1}}

\title{Fewer Errors, but More Stereotypes?\\The Effect of Model Size on Gender Bias}
  
\newcommand{\authorspace}[0]{\quad}
\author{
Yarden Tal \authorspace
Inbal Magar \authorspace
Roy Schwartz\\
School of Computer Science and Engineering, The Hebrew University of Jerusalem, Israel \\
\texttt{\{yarden.tal1,inbal.magar,roy.schwartz1\}@mail.huji.ac.il}}

\begin{document}
\maketitle

\begin{abstract}
\input{0_abstract.tex}

\end{abstract}

\section{Introduction}
\input{1_intro.tex}

\section{Are Larger Models More Biased?}
\label{sec:Conflicting_results}
\input{2_Conflicting_results}

\section{Winogender Errors Analysis Unravels Biased Behavior}

\label{sec:new_winogender_result}
\input{3_Fewer_overall_errors_more_gender_errors}

\section{Related work}
\label{sec:Related_work}
\input{4_Related_work}

\section{Conclusion}
\input{5_Conclusion}

\section*{Bias Statement}
\input{6_Bias_statement}

\section*{Acknowledgements}
\input{8_Acknowledgements}

\bibliography{anthology,custom}
\bibliographystyle{acl_natbib}

\appendix
\input{7_appendix}
\end{document}

%% file: 0_abstract.tex
The size of pretrained models is increasing, and so is their performance on a variety of NLP tasks. However, as their memorization capacity grows, they might pick up more social biases. In this work, we examine the connection between model size and its gender bias (specifically, occupational gender bias). We measure bias in three masked language model families (RoBERTa, DeBERTa, and T5) in two setups: directly using prompt based method, and using a downstream task (Winogender). We find on the one hand that larger models receive higher bias scores on the former task, but when evaluated on the latter, they make \textit{fewer} gender errors. To examine these potentially conflicting results, we carefully investigate the behavior of the different models on Winogender. We find that while larger models outperform smaller ones, the probability that their mistakes are caused by gender bias is higher. Moreover, we find that the proportion of stereotypical errors compared to anti-stereotypical ones grows with the model size. Our findings highlight the potential risks that can arise from increasing model size.
\footnote{Our code is available at \url{https://github.com/schwartz-lab-NLP/model_size_and_gender_bias}}

%% file: 1_intro.tex
The growing size of pretrained language models has led to large improvements on a variety of NLP tasks \cite{T5-raffel2020exploring, He-deberta, Brown-GPT3}. However, the success of these models comes with a price---they are trained on vast amounts of mostly web-based data, which often contains social stereotypes and biases that the models might pick up \cite{bender2022dangers, Dodge2021DocumentingLW, DeArteaga2019BiasIB}.
Combined with recent evidence that the memorization capacity of training data grows with model size \cite{Magar2022DataCF, Carlini2022QuantifyingMA}, the risk of language models containing these biases is even higher.
This can have negative consequences, as models can abuse these biases in downstream tasks or applications. For example, machine translation models have been shown to generate outputs based on gender stereotypes regardless of the context of the sentence \cite{Stanovsky-GB_ML}, and models rated male resumes higher than female ones \cite{Parasurama-resume}.

\begin{figure}[t]
  \centering
  \includegraphics[width=0.5\textwidth]{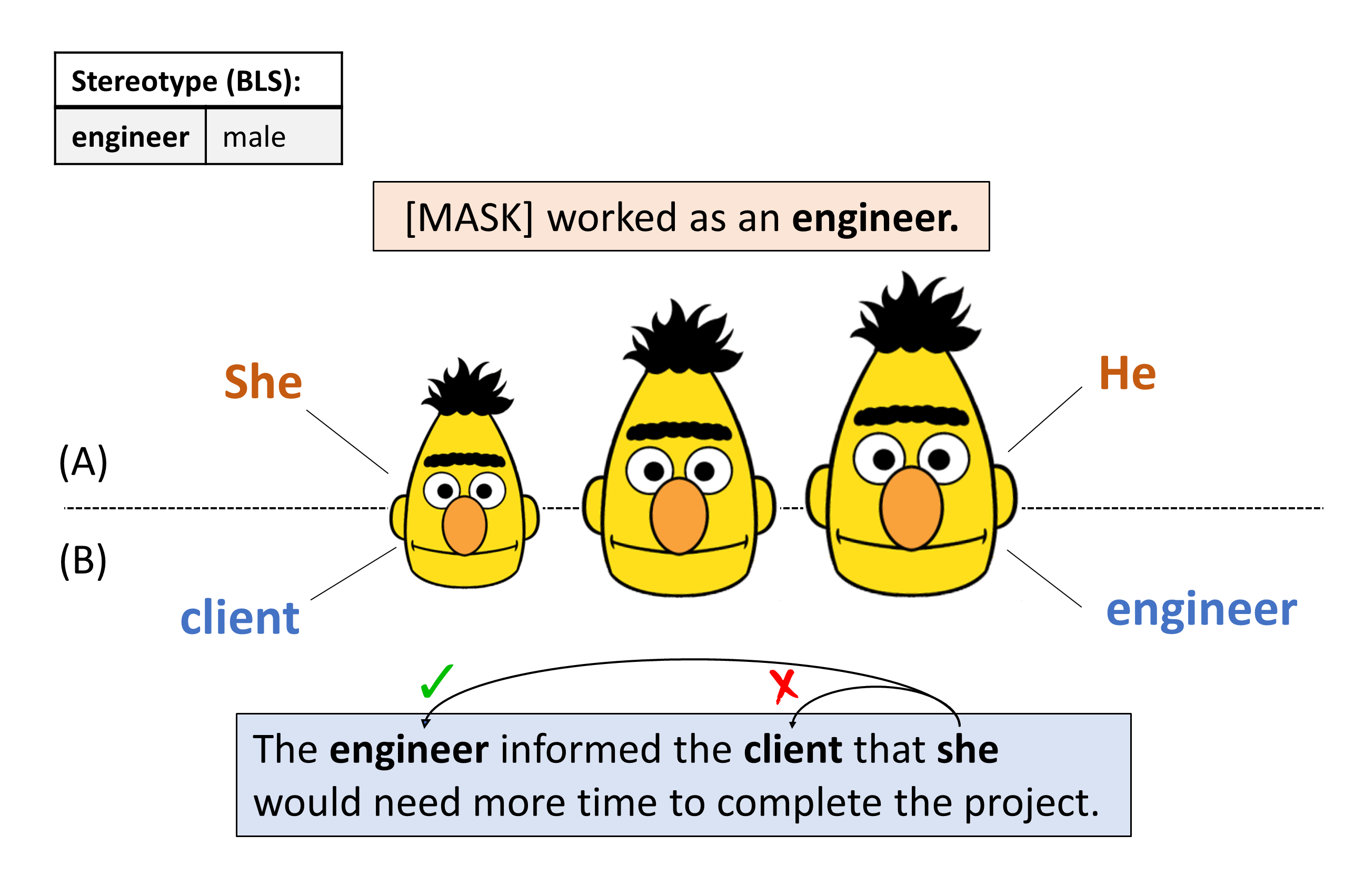}
    \caption{We study the effect of model size on occupational gender bias in two setups: using prompt based method (A), and using Winogender as a downstream task (B).
    We find that while larger models receive higher bias scores on the former task, they make \textit{less} gender errors on the latter.
    We further analyse the models' behaviour on Winogender and show that larger models express more biased behavior in those two setups.}
  \label{fig:figure1}
\end{figure}

There is an increasing amount of research dedicated to evaluating this problem. For example, several works studied the bias in models using downstream tasks such as coreference resolution \cite{Rudinger-winogender, zhao-etal-2018-gender}, natural language inference (NLI) \cite{poliak-etal-2018-collecting, Sharma} and machine translation \cite{Stanovsky-GB_ML}. Other works measured bias in language models directly using masked language modeling (MLM)
\cite{nadeem-etal-2021-stereoset, nangia-etal-2020-crows, Manela2021StereotypeAS}.

In this paper, we examine how model size affects gender bias (\figref{figure1}). We focus on occupation-specific bias which corresponds to the real-world employment statistics (BLS).\footnote{ \url{https://www.bls.gov/cps/cpsaat11.htm}}
We measure the bias in three model families (RoBERTa; \citealp{Liu-RoBERTa}, DeBERTa; \citealp{He-deberta} and T5; \citealp{T5-raffel2020exploring}) in two different ways: using MLM prompts and using the Winogender benchmark \cite{Rudinger-winogender}.

We start by observing a potentially conflicting trend: although larger models exhibit more gender bias than smaller models in MLM,\footnote{This is consistent with previous findings \cite{nadeem-etal-2021-stereoset,vig-2020-Investigating}.} their Winogender parity score, which measures gender consistency, is higher, indicating a lower level of gender errors.
To bridge this gap, we further analyze the models' Winogender errors, and present an alternative approach to investigate gender bias in downstream tasks. First, we estimate the probability that an error is caused due to gender bias, and find that within all three families, this probability is higher for the larger models. Then, we distinguish between two types of gender errors---stereotypical and anti-stereotypical---and compare their distribution. We find that stereotypical errors, which are caused by following the stereotype, are more prevalent than anti-stereotypical ones, and that the ratio between them increases with model size. Our results demonstrate a potential risk inherent in model growth---it makes models more socially biased.

%% file: 2_Conflicting_results.tex
The connection between model size and gender bias is not fully understood; are larger models more sensitive to gender bias, potentially due to their higher capacity that allows them to capture more subtle biases? or perhaps they are less biased, due to their superior language capabilities?

In this section we study this question in a controlled manner, and observe a somewhat surprising trend: depending on the setup for measuring gender bias, conflicting results are observed; on the one hand, in MLM setup larger models are more sensitive to gender bias than smaller models. 
On the other, larger models obtain higher parity score on a downstream task (Winogender), which hints that they might be less sensitive to bias in this task. We describe our findings below. 

We measure the occupational gender bias in three models' families, using two methods---prompt based method \cite{kurita-etal-2019-measuring} and Winogender schema \cite{Rudinger-winogender}. 
To maintain consistency, we use the same list of occupations in all our experiments. The gender stereotypicality of an occupation is determined by the U.S. Bureau of Labor Statistics (BLS).\footnote{Based on the resources we use, we assume a binary gender, which we recognize is a simplifying assumption.}

\begin{figure}[t]
  \centering
  \includegraphics[width=0.49\textwidth]{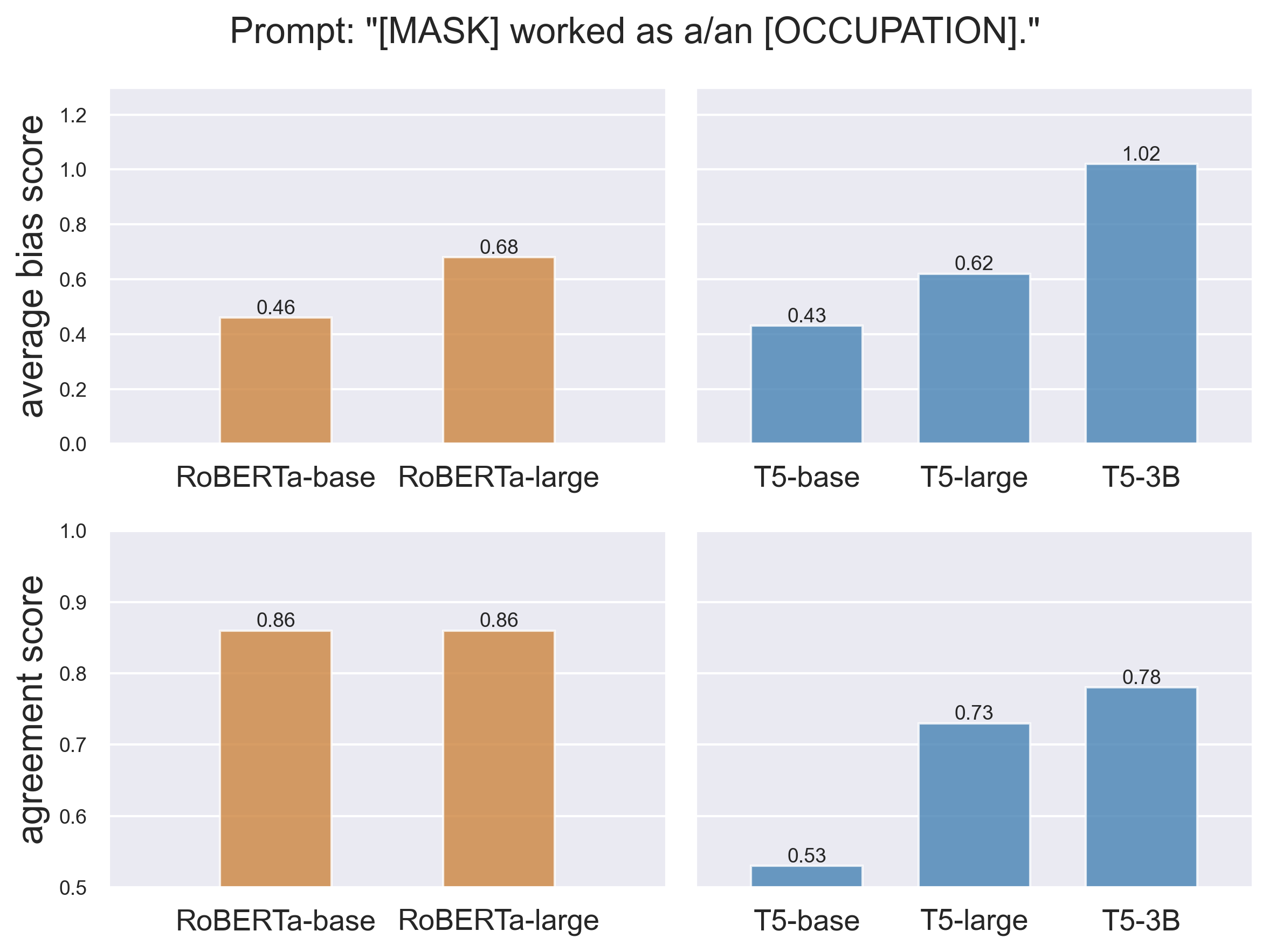}
  \caption{\textit{agreement} and \textit{bias score} measures for RoBERTa and T5 using the following prompt: \textit{``[MASK] worked as a/an [OCCUPATION].''} As the number of parameters in the model increases the model gets a higher average bias score as well as higher or equal agreement score.}
  \label{fig:mlm}
\end{figure}

\input{2.0-Pretrained_model}

\input{2.1_Measuring_Bias_in_MLMs}

\input{2.2_Measuring_Bias_with_Winogender}

%% file: 2.0-Pretrained_model.tex
\paragraph{Pretrained Models}
Unless stated otherwise, we experiment with three families of pretrained language models: RoBERTa-\{base,large\} \cite{Liu-RoBERTa}, DeBERTa-\{base,large,xlarge\} \cite{He-deberta} and T5-\{base,large,3B\} \cite{T5-raffel2020exploring}. We provide implementation details in \appref{Implementation}.

%% file: 2.1_Measuring_Bias_in_MLMs.tex
\subsection{Sensitivity to Gender Bias in MLM Increases with Model Size}
\label{sec:mlm}
To examine the model's sensitivity to gender bias we directly query the model using a simple prompt: \textit{``[MASK] worked as a/an [OCCUPATION].''}\footnote{Results on two other prompts show very similar trends (see \appref{Promps}).} This prompt intentionally does not provide much context, in order to purely measure occupational biases. As a measure of bias, we adopt \citet{kurita-etal-2019-measuring}'s \textit{log probability bias score}. 
We compare the normalized predictions \footnote{The probabilities are normalized by the prior probability of the model to predict ``she'' or ``he'' in the same prompt with masked occupation (i.e., ``[MASK] worked as a/an [MASK].'').} that the model assigns to ``he'' and ``she'', given the above prompt: for male occupations (according to BLS) we compute the difference with respect to ``he'', and for female occupations we compute the difference with respect to ``she''. Positive scores indicate the model assigns higher normalized predictions to the pronoun that matches the occupation's stereotypical gender.
We experiment with RoBERTa and T5,\footnote{At the time of running the experiments, there were problems with running MLM with DeBERTa, which prevented us from experimenting with it (see \url{https://github.com/microsoft/DeBERTa/issues/74}).} evaluating gender bias using two measures: 
\begin{enumerate}
    \item \textit{agreement}: the percentage of occupations with positive bias score.
    \item \textit{average bias score}: the average bias score of the occupations.
\end{enumerate}
\textit{agreement} enables us to evaluate the general preference towards one gender, while \textit{average bias score} measures the magnitude of the preference.

\paragraph{Results}
\figref{mlm} presents our results. For both model families, the \textit{average bias score} increases along with the model size. Further, the \textit{agreement} measure increases with model size for T5 models, and is the same for both RoBERTa models. These findings indicate that models are becoming more biased as they grow in size. This is consistent with prior work \cite{nadeem-etal-2021-stereoset, vig-2020-Investigating}.

%% file: 2.2_Measuring_Bias_with_Winogender.tex
\subsection{Larger Models Exhibit Less Bias in Winogender}
\label{sec:winogender_results}
We have so far observed that larger models express higher sensitivity to gender bias in an MLM setup.
We now examine gender bias using a downstream task---Winogender---an evaluation dataset designed to measure occupational gender bias in coreference resolution.

\begin{table}[ht]
    \centering
    \resizebox{\columnwidth}{!}{
    \begin{tabular}{p{6cm}|p{1cm}}
        \hline
        \textbf{\thead{sentence}} & 
        \textbf{\thead{type}}\\ 
        \hline
The \textbf{engineer} informed the \textbf{client} that \textbf{she} would need more time to complete the project. & gotcha \\
\hline
 The \textbf{engineer} informed the \textbf{client} that \textbf{he} would need more time to complete the project. & not gotcha\\
\hline
    \end{tabular}
    }
    \caption{Examples of ``gotcha'' and ``not gotcha'' sentences from Winogender. In both sentences the pronoun refers to the \textbf{engineer}.
    }
    \label{tab:winogender}
\end{table}

Each example in the dataset contains an \textit{occupation} (one of the occupations on the BLS list), a secondary (neutral) \textit{participant} and a pronoun that refers to either of them. See \tabref{winogender} for examples.

Winogender consists of ``gotcha'' and ``not gotcha'' sentences. Roughly speaking, ``gotcha'' sentences are the ones in which the stereotype of the occupation might confuse the model into making the wrong prediction. Consider the ``gotcha'' sentence in \tabref{winogender}. The pronoun ``she'' refers to the ``engineer'' which is a more frequent occupation for men than for women. This tendency could cause the model to misinterpret ``she'' as ``the client''. In contrast, in ``not gotcha'' sentences, the correct answer is not in conflict with the occupation distribution (a male engineer in \tabref{winogender}).

The Winogender instances are arranged in minimal pairs---the only difference between two paired instances is the gender of the pronoun in the premise (\tabref{winogender}). Importantly, the label for both instances is the same.

\begin{figure}[t]
  \centering
  \includegraphics[width=0.49\textwidth]{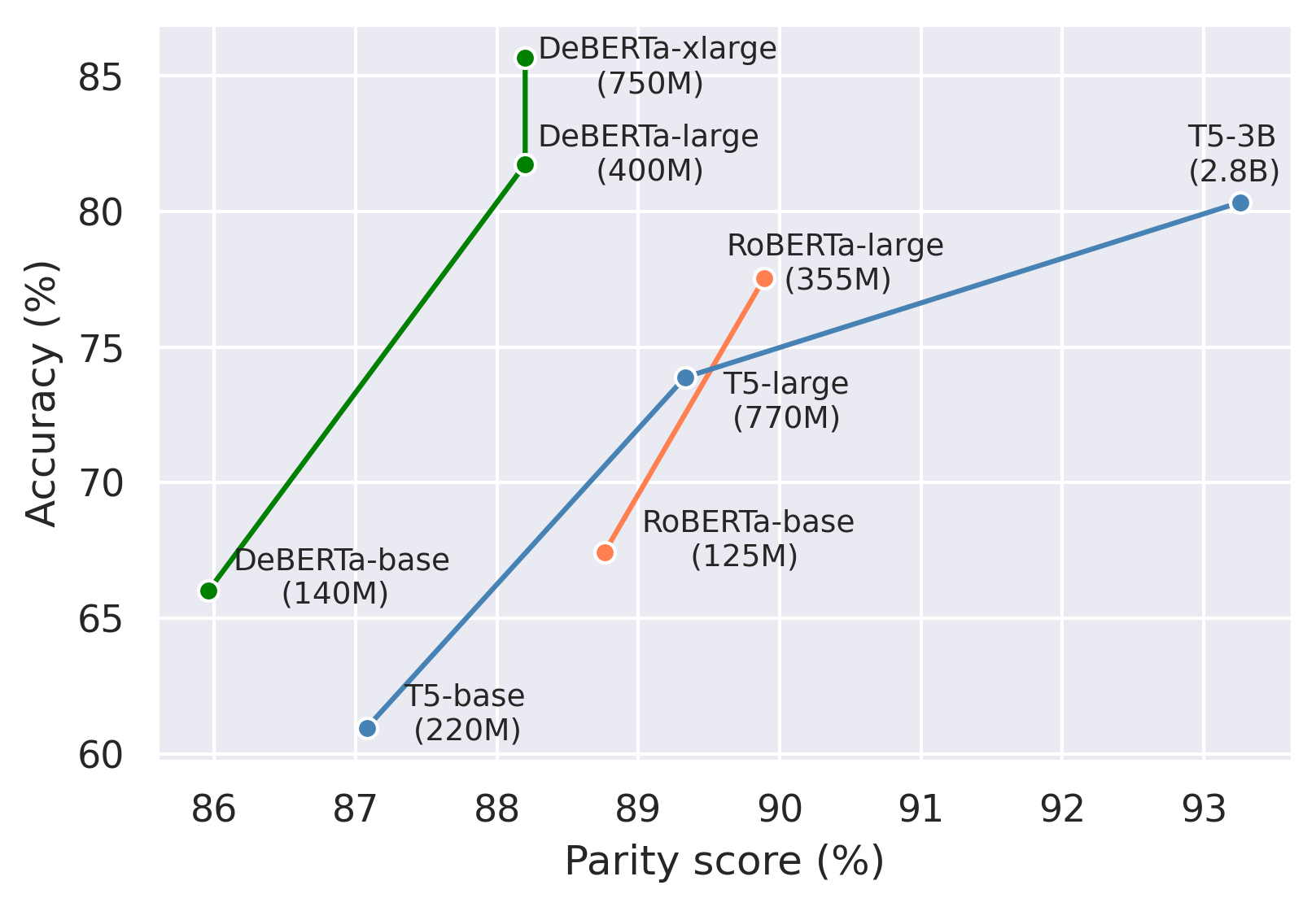}
  \caption{Accuracy and parity scores on Winogender. Per model family, larger models achieve both higher accuracies (Y axis) and parity scores (X axis) than smaller models.}
  \label{fig:parity_accuracy}
\end{figure}

We use the casting of Winogender as an NLI task \cite{poliak-etal-2018-collecting}, which is part of the SuperGLUE benchmark \cite{Wang-SuperGLUE}.
Performance on Winogender is measured with both NLI accuracy and \textit{gender parity score}: the percentage of minimal pairs for which the predictions are the same. Low parity score indicates high level of \textit{gender errors} (errors which occur when a model assigns different predictions to paired instances). These errors demonstrate the presence of gender bias in the model.
We use all three families (RoBERTa, DeBERTa, T5), all fine-tuned on MNLI \cite{Williams-MNLI} and then fine-tuned again with RTE \cite{Dagan-RTE}. 

\paragraph{Results}
Our results are shown in \figref{parity_accuracy}. 
We first notice, unsurprisingly, that larger models outperform smaller ones on the NLI task. Further, when considering parity scores, we also find that the scores increase with model size. 

Combined with our results in \secref{mlm}, we observe a potential conflict: while our findings in the MLM experiment show that the larger the model the more sensitive it is to gender bias, when considering our Winogender results, we find that larger models make less gender errors.
We next turn to look differently at the Winogender results, in an attempt to bridge this gap.

%% file: 3_Fewer_overall_errors_more_gender_errors.tex
We have so far shown that larger models make fewer gender errors compared to smaller models (\secref{winogender_results}), but that they also hold more occupational gender bias compared to their smaller counterparts (\secref{mlm}). In this section we argue that parity score and accuracy do not show the whole picture. Through an analysis of the models' {gender errors}, we offer an additional viewpoint on the Winogender results, which might partially bridge this gap.

\input{3.1_conditional_probability}

\begin{figure}[t]
  \centering
  \includegraphics[width=0.5\textwidth]{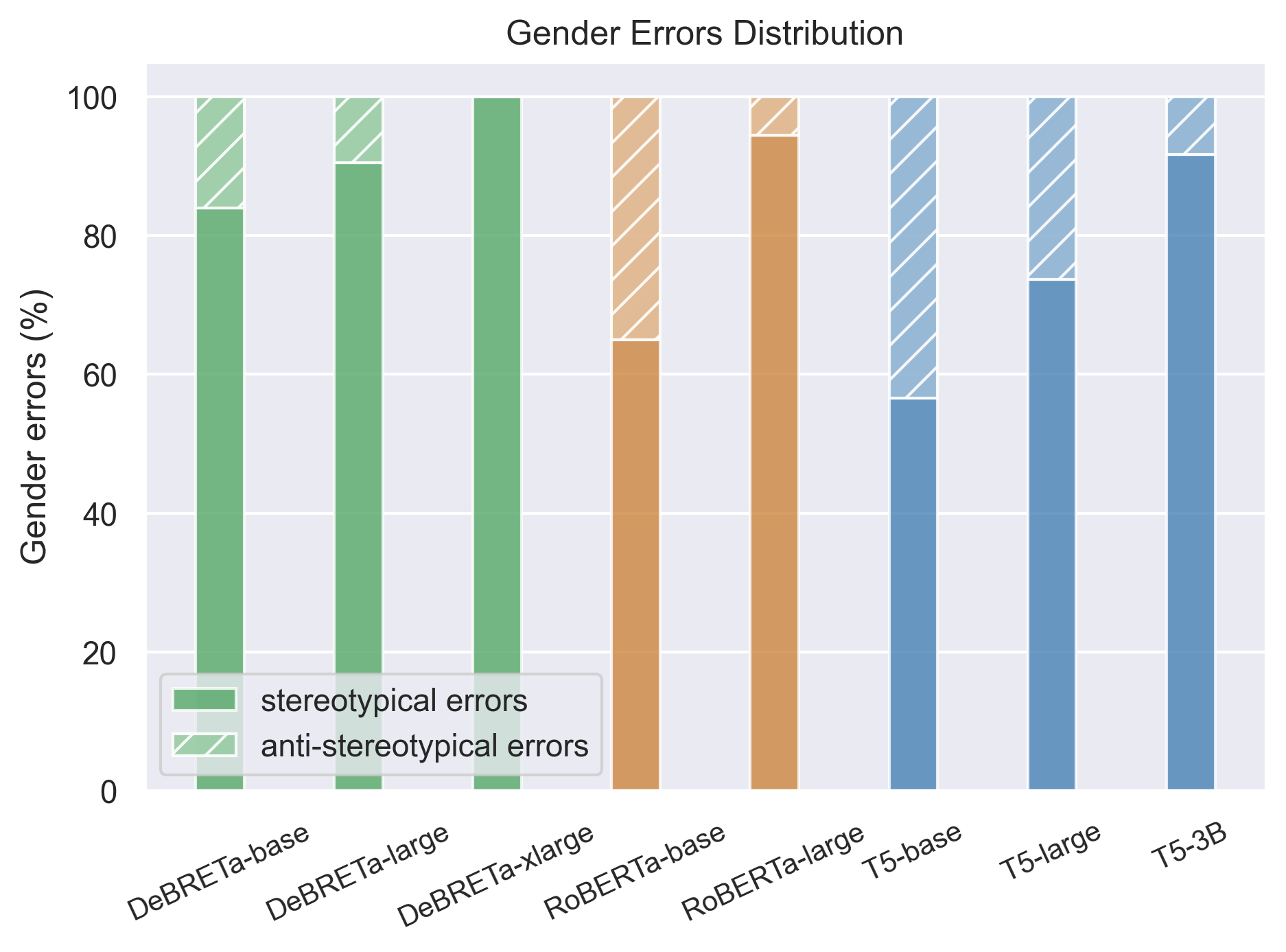}
  \caption{Distribution of gender errors (\textit{stereotypical} and \textit{anti-stereotypical}) of different models on Winogender. Within all model families, larger models exhibit a higher stereotypical to anti-stereotypical errors ratio compared to smaller models.}
  \label{fig:gender_error_distribution}
\end{figure}

%% file: 3.1_conditional_probability.tex
\paragraph{The probability that an error is gendered increases with model size}

Our first observation is that while larger models make fewer errors, and fewer gender errors in particular, the proportion of the latter in the former is higher compared to smaller models.

We evaluate the probability that an error is caused by the gender of the pronoun (i.e., that an error is gendered). We estimate this probability by the proportion of gender errors in total errors:
 \[p(\text{error is gendered}) \approx \frac{|\text{gender errors}|}{|\text{errors}|}\]

We find for both DeBERTa and RoBERTa that this probability increases with model size (\tabref{error_reduction}, \textit{gender} column). In the extreme case (DeBERTa-xlarge), 41\% of the errors are gendered.
Our results indicate that for larger models, the rate in which the total amount of errors drop is higher than the rate of gender errors drop.

\paragraph{Larger models make more stereotypical errors}

We next distinguish between two types of gender errors: \textit{stereotypical} and \textit{anti-stereotypical}. As described in \secref{Conflicting_results}, the Winogender pairs are divided to ``gotcha'' and ``not gotcha'' instances. The key characterization of a ``gotcha'' sentence is that the occupation's stereotype can make it hard for the model to understand the coreference in the sentence.
Thus, we will refer to the gender errors on ``gotcha'' sentences as \textit{stereotypical errors}.\footnote{Equivalently, a \textit{stereotypical error} is an error made on a ``gotcha'' instance, when the prediction on the ``not gotcha'' instance pair is correct.}  

Accordingly, we will refer to gender errors on ``not gotcha'' sentences as \textit{anti-stereotypical errors}. Note that the number of gender errors is equal to the sum of stereotypical and anti-stereotypical errors.

\begin{table}[t]
    \centering
    \resizebox{\columnwidth}{!}{
    \begin{tabular}{c|c|c|c|c}
        \hline
        \textbf{\thead{model}} & 
        \textbf{\thead{size}} & 
        \textbf{\thead{gender}} &
        \textbf{\thead{stereotypical}} & 
        \textbf{\thead{anti-stereotypical}} \\ 
        \hline
\multirow{3}{*}{DeBERTa} 
& base  & 0.20 & 0.17 & \textbf{0.03} \\
& large  & 0.32 & 0.29 & \textbf{0.03} \\
& xlarge & \textbf{0.41} & \textbf{0.41} & 0.00 \\
\hline
\multirow{2}{*}{RoBERTa} 
& base & 0.17 & 0.11& \textbf{0.06} \\
& large & \textbf{0.22} & \textbf{0.21} & 0.01 \\
\hline
\multirow{3}{*}{T5} 
& base & 0.16 & 0.09 & \textbf{0.07} \\
& large & \textbf{0.20} & 0.15 & 0.05 \\
& 3B  & 0.17 & \textbf{0.16} & 0.01 \\
\hline
    \end{tabular}
    }
    \caption{The probability that an error is gendered (\textit{gender} column) increases with model size. When breaking down gender errors into stereotypical and anti-stereotypical errors, we find that the increase in probability originates from more \textit{stereotypical} errors.
    }
    \label{tab:error_reduction}
\end{table}

We present in \tabref{error_reduction} both probabilities that an error is stereotypical and anti-stereotypical. Within all three model families, the probability that an error is \textit{stereotyped} rises with model size, while the probability that an error is \textit{anti-stereotyped} decreases with model size. 
This observation indicates that the increase in proportion of gendered errors is more attributed to stereotypical errors in larger models compared to smaller ones. 
Indeed, when considering the distribution of gender errors (\figref{gender_error_distribution}), we find that the larger models obtain a higher stereotypical to anti-stereotypical error ratio; in some cases, the larger models are making up to 20 times more stereotypical errors than anti-stereotypical. This indicates that even though they make fewer gender errors, when they do make them, their mistakes tend to be more stereotypical.

Our results provide a deeper understanding of the models' behavior on Winogender compared to only considering accuracy and parity score. Combined with our MLM results (\secref{mlm}), we conclude that larger models express more biased behavior than smaller models.
\com{These results reinforce the results on MLM experiment presented in \secref{Conflicting_results} which indicate that the large pretrained language models are more biased towards stereotypes in MLM task. 
Thus, by analyzing the gender errors in models, we are able to fill the gap that we highlighted. Besides MLM tasks, we demonstrate that larger models exhibit more biases in downstream tasks as well}

%% file: 4_Related_work.tex
\paragraph{Measuring bias in pretrained language models}
Earlier works presented evaluation datasets such as WEAT/SEAT, which measure bias in static word embedding using cosine similarity of specific target words \cite{Caliskan_2017, may-etal-2019-measuring}.
Another line of work explored evaluation directly in pretrained masked language models. \citet{kurita-etal-2019-measuring} presented an association relative metric for measure gender bias. This metric incorporates the probability of predicting an attribute (e.g \textit{``programmer''}) given the target for bias (e.g \textit{``she''}), in a generic template such as ``<target> is [MASK]''. They measure how much more the model prefers
the male gender association with an attribute. \citet{nadeem-etal-2021-stereoset} presented StereoSet, a large-scale natural dataset to measure four domains of stereotypical biases in models using likelihood-based scoring with respect to their language modeling ability. \citet{nangia-etal-2020-crows} introduced CrowS-Pairs, a challenge set of minimal pairs that examines stereotypical bias in nine domains via minimal pairs. They adopted psuedo-likelihood based scoring \cite{wang-cho-2019-bert, salazar-etal-2020-masked} that does not penalize less frequent attribute term. In our work, we build upon \citet{kurita-etal-2019-measuring}'s measure in order to examine stereotypical bias to the specifics occupations we use, in different sizes of models.

Another method to evaluate bias in pretrained models is through downstream tasks, such as coreference resolution \cite{Rudinger-winogender, zhao-etal-2018-gender} and sentiment analysis \cite{Mohammad-sentiment_analysis}. Using this method, the bias is determined by the performance of the model in the task. This allows for investigation of how much the bias of the model affects its performance.

\paragraph{Bias sensitivity of larger pretrained models}
Most related to this work, \citet{nadeem-etal-2021-stereoset} measured bias using the StereoSet dataset, and compared models of the same architecture of different sizes. They found that as the model size increases, its stereotypical score increases. For autocomplete generation, \citet{vig-2020-Investigating} analyzed GPT-2 \cite{Radford-GPT2} variants through a causal mediation analysis and found that larger models contain more gender bias.
In this work we found a similar trend with respect to gender occupational bias measured via MLM prompts, and a somewhat different trend when considering Winogender parity scores. Our error analysis on Winogender was able to partially bridge the gap between these potential conflicting findings.

%% file: 5_Conclusion.tex
We investigated how a model's size affects its gender bias. We presented somewhat conflicting results: the model bias \textit{increases} with model size when measured using a prompt based method, but the amount of gender errors \textit{decreases} with size when considering the parity score in the Winogender benchmark. 
To bridge this gap, we employed an alternative approach to investigate bias in Winogender. Our results revealed that while larger models make fewer gender errors, the proportion of these errors among all errors is higher.
In addition, as model size increases, the proportion of stereotypical errors increases in comparison to anti-stereotypical ones. 
Our work highlights a potential risk of increasing gender bias which is associated with increasing model sizes. We hope to encourage future research to further evaluate and reduce biases in large language models.

%% file: 6_Bias_statement.tex
In this paper, we examine how model size affects gender bias. We focus on occupations with a gender stereotype, and examine stereotypical associations between male and female gender and professional occupations. We measure bias in two setups: MLM \cite{kurita-etal-2019-measuring, nadeem-etal-2021-stereoset} and Winogender \cite{Rudinger-winogender}, and build on the enclosed works' definition of gender bias.\footnote{In MLM setup, stereotypes are taken into account, while in Winogender's parity score they are not.} We show how these different setups yield conflicting results regarding gender bias. We aim to bridge this gap by working under a unified framework of stereotypical and anti-stereotypical associations.
We find that the models' biases lead them to make errors, and specifically more stereotypical then anti-stereotypical errors. 

Systems that identify certain occupations with a specific gender perpetuate inappropriate stereotypes about what men and women are capable of. Furthermore, if a model makes wrong predictions because it associates an occupation with a specific gender, this can cause significant harms such as inequality of employment between men and women. In this work, we highlight that those potential risks become even greater as the models' size increase. Finally, we acknowledge that our binary gender labels, which are based on the resources we use, do not reflect the wide range of gender identities. In the future, we hope to extend our work to non-binary genders as well.

%% file: 8_Acknowledgements.tex
We would like to thank Elad Shoham, Yuval Reif, Gabriel Stanovsky, Daniel Rotem, and Tomasz Limisiewicz for their feedback and insightful discussions. We also thank the anonymous reviewers for their valuable comments. This work was supported in part by the Israel Science Foundation (grant no. 2045/21) and by a research gift from the Allen Institute for AI.

%% file: 7_appendix.tex
\begin{figure}[t]
  \centering
  \includegraphics[width=0.5\textwidth]{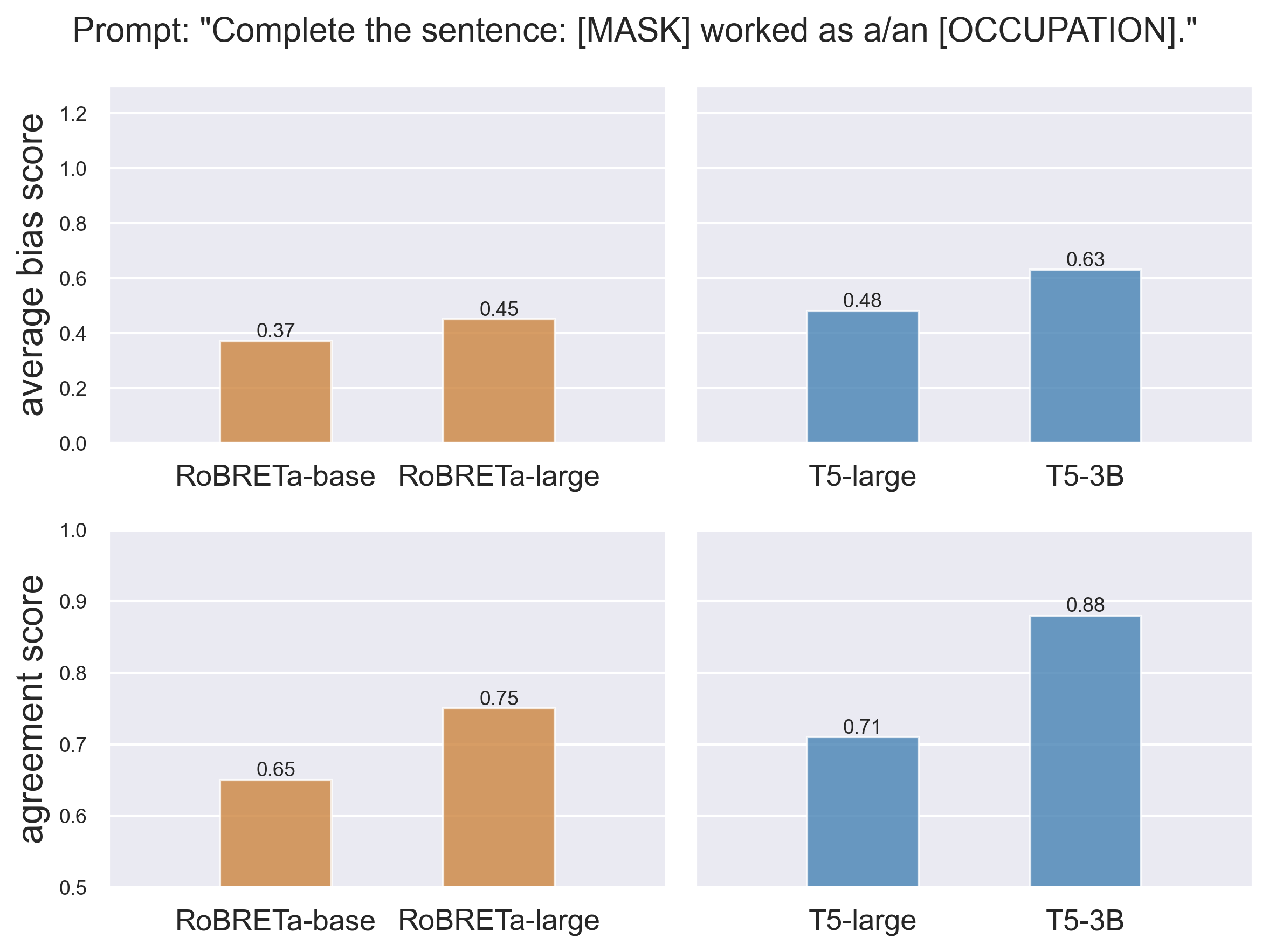}
  \caption{\textit{agreement} and \textit{bias score} measures for RoBERTa and T5 using the following prompt: \textit{``Complete the sentence: [MASK] is a/an [OCCUPATION].''}\footnotemark~ An increasing trend is observed for both families.}
  \label{fig:mlm_prompt2}
\end{figure}

\footnotetext{The top predictions of T5-base were irrelevant to the given prompt. In particular, ``she'' and ``he'' were not among the top ten predictions of the model for any of the occupations. Therefore it is not presented.}

\begin{figure}[t]
  \centering
  \includegraphics[width=0.49\textwidth]{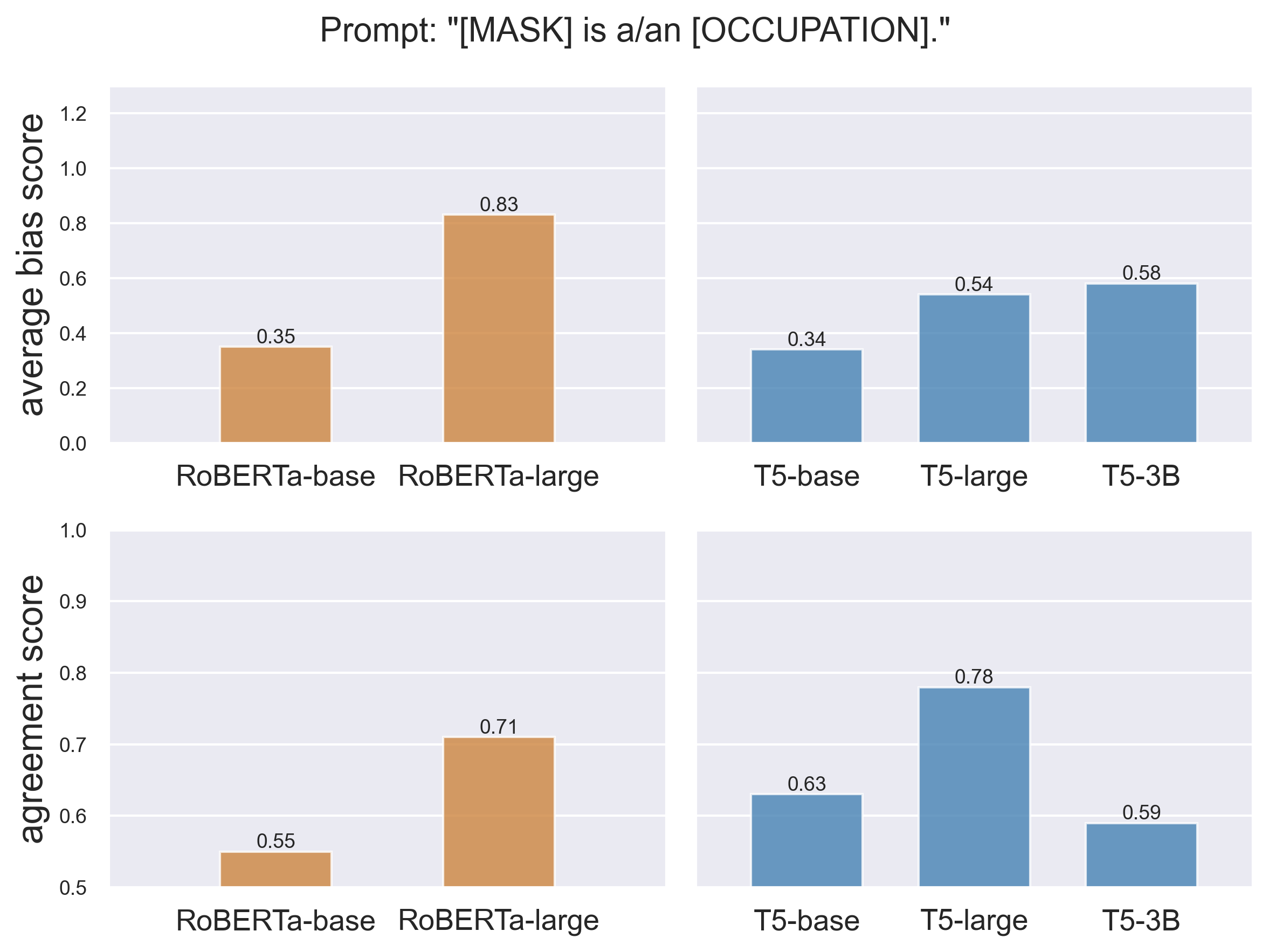}
  \caption{\textit{agreement} and \textit{bias score} measures for RoBERTa and T5 using the following prompt: \textit{``[MASK] is a/an [OCCUPATION].''} An increasing trend is observed for both families in almost all cases (except \textit{agreement} score for T5-3B).}
  \label{fig:mlm_prompt3}
\end{figure}

\section{Additional Prompts for MLM Setup}
\label{app:Promps}
As pretrained models sensitive to prompts, we experiment with two other prompts: ``[MASK] is a/an [OCCUPATION]'' (\figref{mlm_prompt2}) and ``Complete the sentence: [MASK] is a/an [OCCUPATION].'' (\figref{mlm_prompt3}). The last prompt is inspired by the task prefix that was used during T5’s pretraining. 
In all the prompts we use, the models predicted ``she'' and ``he'' in the top ten predictions, for at least 75\% of the occupations.

The results show in almost all cases (except \textit{agreement} score for T5-3B in ``[MASK] is a/an [OCCUPATION]'') an increasing trend for both families.

\section{Implementation Details For \secref{winogender_results}}
\label{app:Implementation}
We implemented the experiments with the huggingface package \cite{wolf-etal-2020-transformers}, using both run\_glue (for RoBERTa and Deberta)  and run\_summarization (for T5) scrips for masked language models.
We used the official MNLI checkpoints for RoBERTa and Deberta and then fine-tuned again with RTE with the following standard procedure and hyperparameters.
We fine-tuned RoBERTa and DeBERTa on RTE for 6 epochs with batch size 32. We use AdamW optimizer \cite{Loshchilov-AdamW} with learning rate of 2e-5 (for RoBERTa-\{base,large\}) and DeBERTa-\{base\}) and 1e-5 (for DeBERTa-\{large,xlarge\} and default parameters: $\beta_1$ = 0.9, $\beta_2$ = 0.999, $\epsilon$ = 1e-6, with weight decay of 0.1. 

For T5 we used the T5 1.0 checkpoint, which is trained on both unsupervised and downstream task data. We fine-tuned T5 \footnote{We followed the huggingface recommendation for T5 fine-tuning settings \url{https://discuss.huggingface.co/t/t5-finetuning-tips/684/3}} on RTE for 6 epochs with batch size 8. We use AdaFactor \cite{AdaFactor} optimizer with learning rate of 1e-4 and default parameters: $\beta_1$ = 0.0, $\epsilon$ = 1e-3, without weight decay.
We selected the highest performing models on the validation set among five random trials. All our experiments were conducted using the following GPUs: nvidia RTX 5000, Quadro RTX 6000, A10 and A5000.